\documentclass[11pt,a4paper]{article}
\usepackage[hyperref]{emnlp2020}
\usepackage{times}
\usepackage{latexsym}

\usepackage{amsmath}
\usepackage{amssymb}
\usepackage{amsthm}
\usepackage{amscd}
\usepackage{mathrsfs}  
\usepackage{calrsfs}
\usepackage{bm}
\usepackage{graphicx}

\usepackage{microtype}

\aclfinalcopy 
\def\aclpaperid{2476} 

\usepackage{booktabs}
\usepackage{xspace}
\usepackage{tikz}
\usepackage{xcolor}
\usepackage{graphics}
\usepackage{multirow}
\usepackage{multicol}
\usepackage{lipsum}  

\newcommand{\rarrow}{%
\parbox{0.3cm}{\centering\tikz{\draw[->](0cm,0)--(0.2cm,0);}}
}

\newcommand{\minus}{%
\parbox{0.15cm}{\centering\tikz{\draw[-](0cm,0)--(0.15cm,0);}}
}

\newcommand{\Tab}[1]{Tab.~\ref{#1}}
\newcommand{\Fig}[1]{Fig.~\ref{#1}}

\newcommand{\Sec}[1]{Section~\ref{#1}}

\newcommand{\tAA}{\ensuremath{T_{\textsc{aa}}}\xspace}
\newcommand{\tBB}{\ensuremath{T_{\textsc{bb}}}\xspace}
\newcommand{\tAB}{\ensuremath{T_{\textsc{ab}}}\xspace}
\newcommand{\tBA}{\ensuremath{T_{\textsc{ba}}}\xspace}
\newcommand{\tABA}{\ensuremath{T_{\textsc{aba}}}\xspace}
\newcommand{\tBAB}{\ensuremath{T_{\textsc{bab}}}\xspace}
\newcommand{\tBmB}{\ensuremath{T_{\textsc{b}^{m}\textsc{b}}}\xspace}

\newcommand{\xS}{\ensuremath{x_{\textsc{s}}}\xspace}
\newcommand{\xP}{\ensuremath{x_{\textsc{p}}}\xspace}

\newcommand{\xA}{\ensuremath{x_{\textsc{a}}}\xspace}
\newcommand{\xB}{\ensuremath{x_{\textsc{b}}}\xspace}

\newcommand{\hxB}{\ensuremath{\hat{x}_{\textsc{b}}}\xspace}

\newcommand{\xAA}{\ensuremath{x_{\textsc{aa}}}\xspace}
\newcommand{\xBB}{\ensuremath{x_{\textsc{bb}}}\xspace}
\newcommand{\xAB}{\ensuremath{x_{\textsc{ab}}}\xspace}
\newcommand{\xBA}{\ensuremath{x_{\textsc{ba}}}\xspace}

\newcommand{\xABA}{\ensuremath{x_{\textsc{aba}}}\xspace}
\newcommand{\xBAB}{\ensuremath{x_{\textsc{bab}}}\xspace}

\newcommand{\Lrec}{\ensuremath{\mathcal{L}_{\textsc{rec}}}\xspace}
\newcommand{\Lsup}{\ensuremath{\mathcal{L}_{\textsc{sup}}}\xspace}
\newcommand{\Lbt}{\ensuremath{\mathcal{L}_{\textsc{bt}}}\xspace}

\newcommand{\sep}{\textsc{[sep]}\xspace}
\newcommand{\mask}{\textsc{[mask]}\xspace}

\title{DualTKB: A Dual Learning Bridge between Text and Knowledge Base}

\author{Pierre L. Dognin\thanks{* Equal contribution} \\ IBM Research \\ \texttt{pdognin@us.ibm.com} \And
  Igor Melnyk\textsuperscript{*} \\ IBM Research \\ \texttt{igor.melnyk@ibm.com} \And
  Inkit Padhi\textsuperscript{*} \\ IBM Research \\ \texttt{inkpad@ibm.com} \\ \AND
  Cicero Nogueira dos Santos \\ AWS AI \\ \texttt{cicnog@amazon.com} \And
  Payel Das\\IBM Research \\ \texttt{daspa@us.ibm.com}}

\date{}

\begin{document}
\maketitle

\begin{abstract}
In this work, we present a dual learning approach for unsupervised text to path and path to text transfers in Commonsense Knowledge Bases (KBs). We investigate the impact of weak supervision by creating a weakly supervised dataset and show that even a slight amount of supervision can significantly improve the model performance and enable better-quality transfers. We examine different model architectures, and evaluation metrics, proposing a novel Commonsense KB completion metric tailored for generative models.
Extensive experimental results show that the proposed method compares very favorably to the existing baselines. This approach is a viable step towards a more advanced system for automatic KB construction/expansion and the reverse operation of KB conversion to coherent textual descriptions.

\end{abstract}

\section{Introduction}
\label{sec:intro}

The automatic construction of Knowledge Bases (KBs) from text and the reverse operation of sentence generation from KBs are dual tasks that are both active research topics.
\begin{figure}[!t]
\centering
\includegraphics[width=\columnwidth]{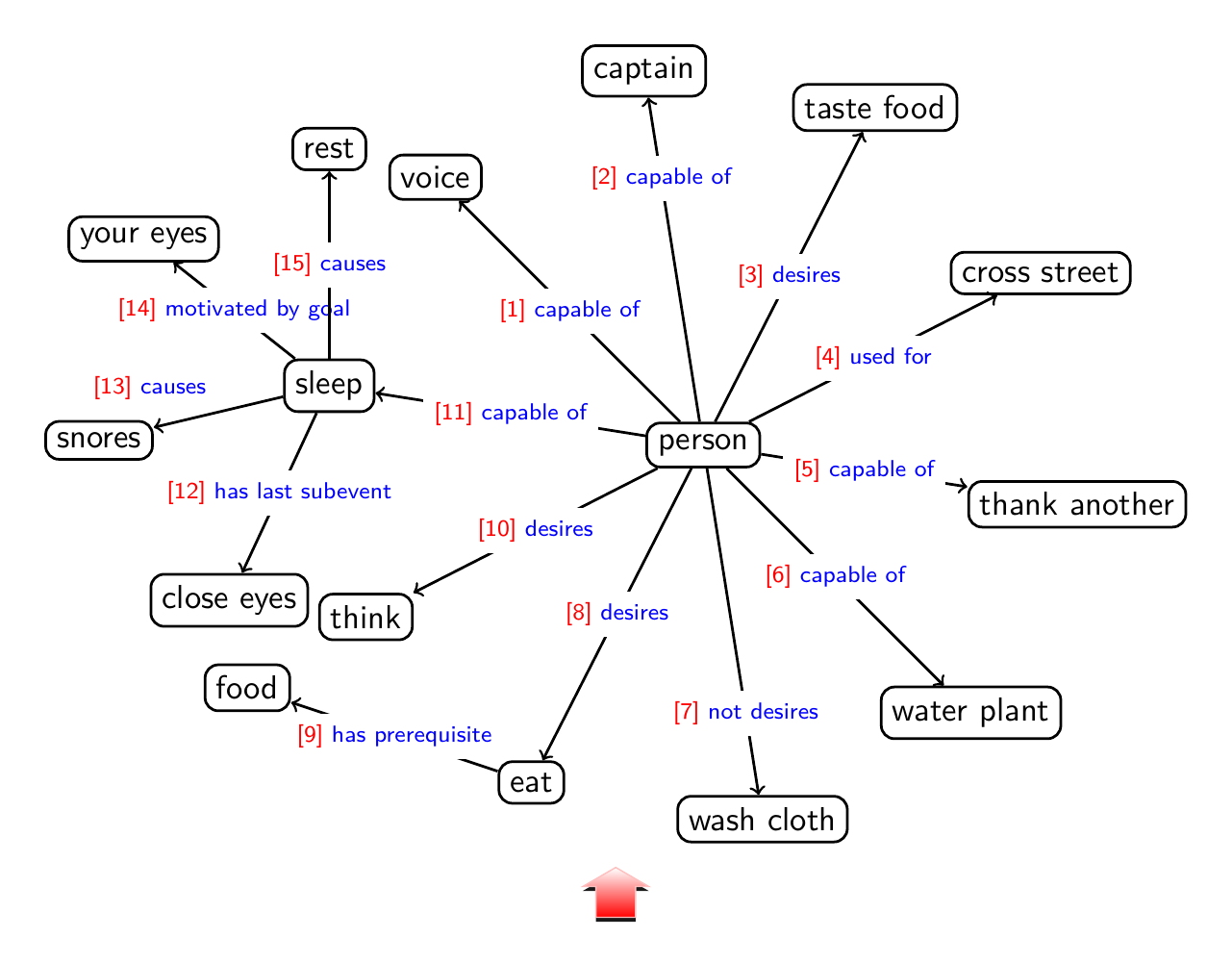}
{
\centering
\resizebox{\columnwidth}{!}{
\begin{tabular}{clcl}
    & & & \\
    {[}1{]} & a person can voice an opinion &
    {[}2{]} & a person can captain a ship \\
    {[}3{]} & a person can taste food &
    {[}4{]} & person can cross the street\\
    {[}5{]} & person can thank another person &
    {[}6{]} & a person can water a plant \\
    {[}7{]} & person can wash cloths & 
    {[}8{]} & person can eat\\
    {[}9{]} & you eat food & 
    {[}10{]} & a person can think\\
    {[}11{]} & a person can sleep & & \\
    {[}12{]} & \multicolumn{3}{l}{if you want to sleep then you should close eyes}\\
    {[}13{]} & \multicolumn{3}{l}{sleep would make you want to snore} \\
    {[}14{]} & \multicolumn{3}{l}{if you want to sleep then you should close your eyes} \\
    {[}15{]} & you sleep to rest.  & & \\
\end{tabular}
}
}
\caption{Text to Path. Part of a larger graph generated from test sentences from our dataset. Sentences below the graph were a subset of inputs provided to the model.}
\label{fig:gengraph}
\end{figure}

The first task of automatic KB construction remains a significant challenge due to the difficulty of detecting parts of text representing meaningful facts and summarizing them in a systematic form. A simpler sub-task of KB completion, i.e., extending or filling-in missing nodes or edges, has also attracted the attention of the research community. For both tasks, the system needs to generate new or complete existing graph entities coherently, possibly matching to the already existing graph structure. 
The dual task of decoding the information from KB back to text is a valuable functionality. This enables knowledge transfer from potentially large complex graphs into a more descriptive, human-friendly output. This conditional generation is often seen as a step towards learning using KB as prior knowledge.

In this work, we address the problem of KB construction/completion and the reverse task of KB decoding, but aim at a simpler objective:  transferring a single sentence to a path, and generating text from a single KB path as its dual task.  
In terms of data, our focus will be on Commonsense KBs, derived from sets of commonsense facts expressed in natural language sentences  \citep{lenat1995cyc, cambria2014senticnet, speer2017conceptnet, sap2019atomic}. They are represented as graphs where each edge is expressed as a tuple $(e_h,r,e_t)$ with head and tail nodes $e_h$ and $e_t$ composed of free-form text, connected with a relationship operator $r$; see ConceptNet from \citet{speer2017conceptnet} or ATOMIC in \citealp{sap2019atomic} for recent and commonly used examples of commonsense KBs.

We observe that to train such transfer model, an additional challenge comes from the lack of datasets with parallel text and KB facts, i.e., where text sentences and KBs edges are explicitly paired/labeled from one to another. However, there exist many datasets for each individual transfer domain. Therefore, successful approaches transferring text to KB and KB to text must be able to operate in unsupervised or (at best) weakly-supervised settings. 
We address this challenge by proposing a model trained under dual learning of translation/transfer from text to KB and from KB to text, we name \emph{DualTKB}. This is similar in philosophy to dual learning in Neural Machine Translation \citep{dual_nips2016}, or unsupervised style transfer \citep{shen2017style, tian2018structured, dai2019style}.
We design our model to be trained in completely unsupervised settings. However, we observed that even a slight supervision significantly boosts model performance and enables better-quality transfers. Therefore, we also describe a simple heuristic methodology to create weakly-supervised datasets given a text corpus and a commonsense KB.

We must emphasize that our proposed dual learning method is not limited to commonsense KBs and can generalize to other domains/types of KBs  such as biomedical KBs. Commonsense KBs, and particularly ConceptNet, are good starting points due to the nature of their composition. 
Since ConceptNet was partly extracted from free-form text originating from the Open Mind Common Sense (OMCS) list of commonsense fact sentences, its nodes are often composed of parts of sentences from OMCS. 
This allowed us to first explore whether the proposed method worked at all before evaluating a semi-supervised approach by creating a weak supervision from a mapping between ConceptNet triples and the original OMCS sentences. 
While KBs are often dense with short named entity descriptions for nodes, many nodes for commonsense KBs are parts of sentences, making them inherently sparse which impacts their performance as empirically studied by \citet{malaviya2020commonsense}.

\begin{figure}[!t]
\centering
\includegraphics[width=\columnwidth]{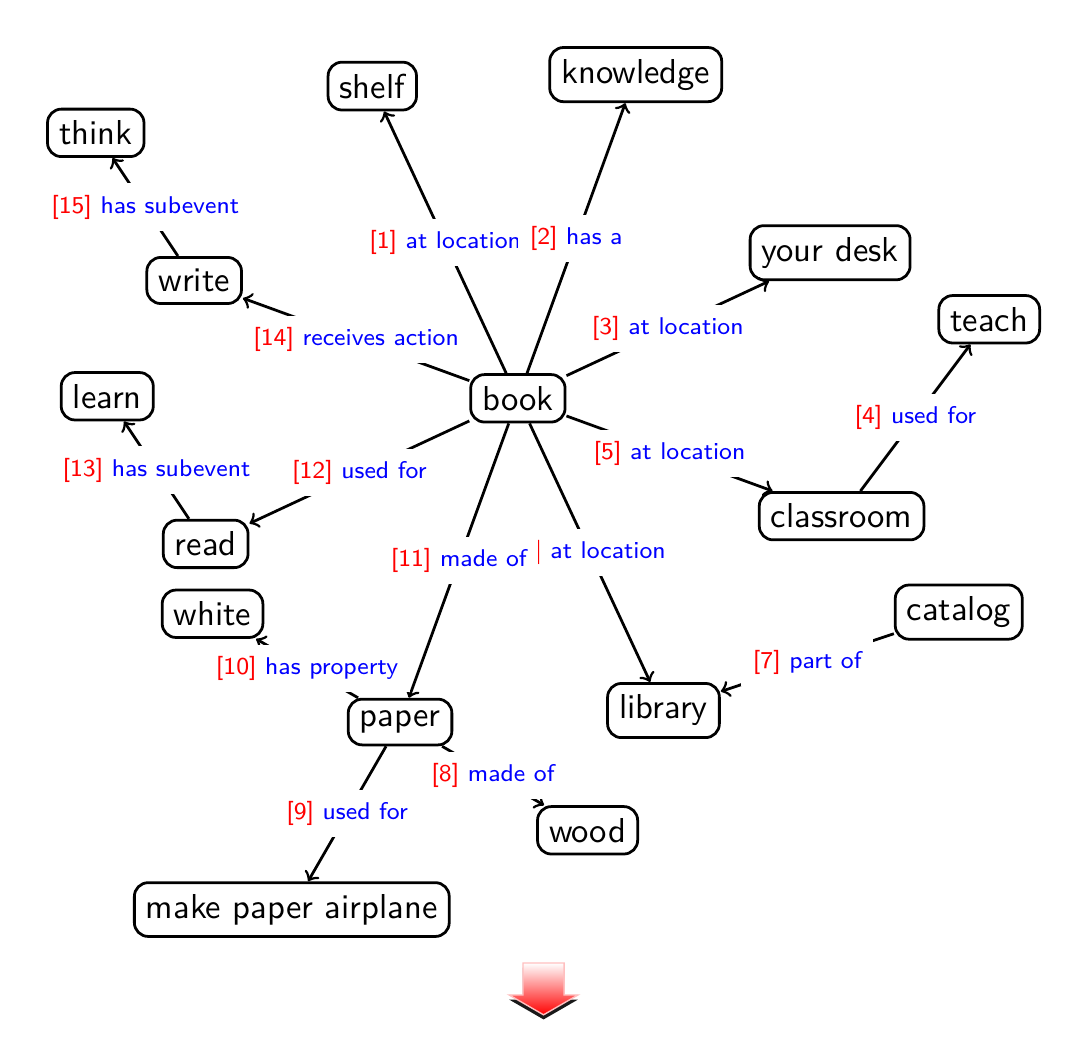}
{
\centering
\resizebox{\columnwidth}{!}{
\begin{tabular}{clcl}
    & & & \\
    {[}1{]} &  something you find on a shelf is a book &
    {[}2{]} & a book is have knowledge  \\
    {[}3{]} & something you find to find desk is a book &
    {[}4{]} & one can use a classroom to teach  \\
    {[}5{]} & you are likely to find a book in a classroom &
    {[}6{]} & something can be at the library \\
    {[}7{]} & you are likely to find a card catalog in a library  & 
    {[}8{]} & paper is be made of wood \\
    {[}9{]} & paper is is for making paper airplane  & 
    {[}10{]} & paper is white \\
    {[}11{]} & a book is a book of paper  & 
    {[}12{]} & one can read a book \\
    {[}13{]} & you can read to learn  & 
    {[}14{]} & you can write a book \\
    {[}15{]} & something you thing you do when you write is think  & 
\end{tabular}
}
}
\caption{Path to Text. Sentences generated by our system from a subgrapth of the ConceptNet dataset. The paths shown in the graph are the inputs to the model.}
\label{fig:gentext}
\end{figure}

The evaluation of this type of transfer models is a challenge in itself. For this purpose, we selected a set of metrics to examine different facets of the system using our created weakly-supervised dataset. For path generation, we rely on a \emph{conventional} KB completion task where the goal is to maximize the validity score of a tail entity $e_t$ given the pair $(e_h,r)$. For example, \citet{malaviya2020commonsense} addresses the challenges unique to commonsense KB completion due to sparsity and large numbers of nodes resulting from encoding commonsense facts.
However, KB completion does not always equate generation of edges, with the exception of COMET from \citet{bosselut2019comet} that generates tail node $e_t$ given the pair $(e_h,r)$.

Since repurposing generative models for conventional KB completion evaluation is difficult \citep{malaviya2020commonsense, bosselut2019comet}, we propose a new commonsense KB completion evaluation task for generative models. It is close in spirit to conventional KB completion, but comes with its own set of challenges. Moreover, we employ the Graph Edit Distance (GED) to examine the quality of the generated graph as a whole. For text generation, we rely on traditional NLP metrics such as BLEU and ROUGE.

Following is a list of highlights of our paper contributions: (1) Propose a dual learning bridge between text and commonsense KB. Implement approach as unsupervised text-to-path and path-to-text transfers; (2) Construct a weakly-supervised dataset, and explore weak-supervision training. (3) Define a novel Commonsense KB completion metric tailored for generative models. (4) Investigate successfully multiple model architectures.

Finally, in \Fig{fig:gengraph} and \Fig{fig:gentext} we present a few examples generated by our proposed model. \Fig{fig:gengraph} is a text to KB translation. Each sentence below the graph is independently transferred to a path, consisting of one edge tuple $(e_h,r,e_t)$. The whole path tuple is generated at once, with $e_h$, $r$, and $e_t$ taking a free-form not restricted to any predefined sets of entities or relations. 
This contrasts with many existing works operating on a limited discrete set of already-defined edges in a dense conventional KB. Once all the sentences are transferred, we observe that this set of generated edges forms a connected structure, implicitly merging some nodes to form a prototype of a Knowledge Graph (KG). 

\Fig{fig:gentext} shows the transfer from KB to text. Each path in the graph is converted to a sentence. There is overall diversity in the generated sentences styles. Moreover, the samples show that the generation process is more sophisticated than just a trivial path flattening (i.e., merging text from all edge parts followed by minimal edits). Therefore, the proposed approach can eventually become a part of a more sophisticated system converting graphs to a coherent textual story and vice versa. Additional examples are presented in Appendix \ref{sec:app_graph_gen}.

\section{Dual Learning}
\label{sec:approach}

\begin{figure*}[th]
\centering
\includegraphics[width=\textwidth]{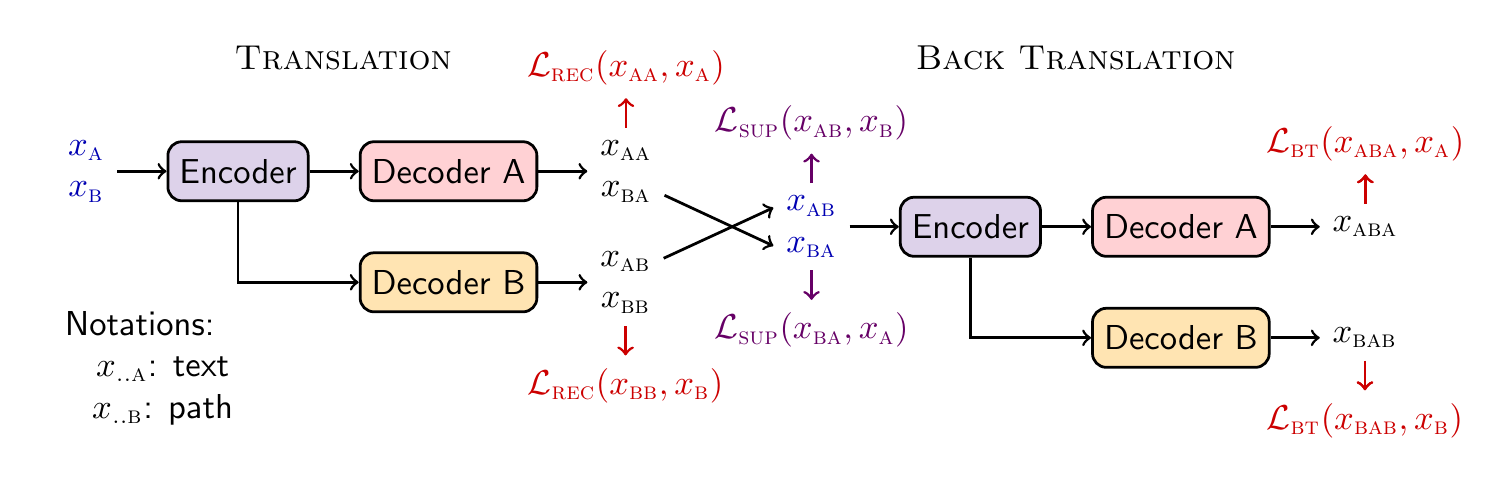}
\caption{Our Encoder-Decoder model with translation and back-translation represented as a sequence of steps. Encoder and decoders are shared for both steps. The Encoder provides a representation of inputs \xA (text) and \xB (path) in a common embedding space. Decoders A and B are specialized to generate only sentences and paths (respectively) from these embeddings. Losses are indicated close to the generated values they require. Reconstruction loss \Lrec, back-translation loss \Lbt are available in unsupervised learning, with \Lsup only available in supervised learning. Generation of \xABA from \xA means two passes through our model: First pass, with \xA as input. Second pass, with $\xAB=\tAB(\xA)$. The same model can accommodate path or text generation.}
\label{fig:model_simple}
\end{figure*}

In this work we propose to use a dual learning approach to build a model performing two distinct but complementary generative tasks described in \Fig{fig:model_simple}, and using notations from \Tab{tab:notations}:

\noindent\textbf{Task 1 ($\text{text}\rarrow\text{path}$):}
Given a sentence \xA, generate a path \xAB with well-formed entities and relation, that can either belong to an already constructed KB, or extend it in a factually meaningful way. 
This conditional generation is framed as a translation task referred to as $\tAB$ where $\xAB=\tAB(\xA)$.

\noindent\textbf{Task 2 ($\text{path}\rarrow\text{text}$):}
Given a KB path \xB, generate a descriptive sentence \xA, coherently merging entities and relation from the path.
This conditional generation is a translation task referred to as $\tBA$, where $\xBA=\tBA(\xB)$.

From the above two tasks follows the definition of \emph{back-translation} tasks \tBAB ($\text{path}\rarrow\text{text} \rarrow\text{path}$) and \tABA ($\text{text}\rarrow\text{path} \rarrow\text{text}$). Reconstruction tasks \tAA and \tBB are trivially defined as generating the same text/path from itself. 

In an unsupervised setting, where sentences \xA and paths \xB are not paired, the reconstruction tasks \tAA, \tBB and back-translation tasks \tABA, \tBAB are the only ones available to define training losses. 
The back-translation tasks define the so-called cycle/consistency losses, which implicitly control the quality of the first transfers (\tAB, \tBA), by checking the reconstruction after the second transfers (\tABA, \tBAB). By themselves, these cycle losses are effective in training good transfer model. However, as we show in \Sec{sec:expts}, even a small weak supervision (by pairing some \xA to \xB) can significantly improve our model performance.

As shown in \Fig{fig:model_simple}, our model uses an encoder-decoder architecture with a common encoder and two specialized decoders (A, generating sentences \xA, and B, generating paths \xB). The reason for a single encoder is to force the embeddings of path and text to lie in the same space, encoding a fact regardless of modality. It then becomes the job of each specific decoder to recover either one of the modalities. Note that the decoders share the same architecture, but not their parameters.



\begin{table}[t]
\centering
\resizebox{\columnwidth}{!}{
\begin{tabular}{lll}
\toprule
\xS & \multicolumn{2}{l}{sentence, where type \textsc{s} can be} \\
    & A, A$^m$ & sentence, masked sentence \\
    & AA       & given A, reconstructed sentence\\
    & BA       & given B, generated sentence\\
    & ABA      & given AB, back-translated sentence\\
\midrule
\xP & \multicolumn{2}{l}{path, where type \textsc{p} can be} \\
    & B, B$^m$  & path, masked path\\
    & BB        &  given B, reconstructed path\\
    & B$^m$B    &  given B$^m$, reconstructed path\\
    & AB        &  given A, generated path\\
    & BAB       &  given BA, back-translated B\\
$w$ & \multicolumn{2}{l}{edge $w=(e_h, r, e_t)$ with (entity, rel. op., entity)} \\
$w^m$ & \multicolumn{2}{l}{masked edge/tuple, e.g., $(e_h, r, .)$, $(., r, e_t)$, etc.} \\
\midrule
$T_\textsc{d}$ & \multicolumn{2}{l}{Translation task with direction \textsc{d}}  \\
 \tAB & \multicolumn{2}{l}{Translation from A to B. $\xAB=\tAB(\xA)$} \\
 \tBA & \multicolumn{2}{l}{Translation from B to A. $\xBA=\tBA(\xB)$} \\
 \tABA & \multicolumn{2}{l}{Back-Translation from AB, back to A}\\
       & \multicolumn{2}{l}{where $\xABA=\tBA(\xAB)=\tBA(\tAB(\xA))$}\\
 \tBAB & \multicolumn{2}{l}{Back-Translation from BA to B}\\
    & \multicolumn{2}{l}{where $\xBAB=\tAB(\xBA)=\tAB(\tBA(\xB))$} \\
\tBmB & \multicolumn{2}{l}{Generation from masked path $\xP=\tBmB(\xP^m)$}\\
\bottomrule
\end{tabular}
}
\caption{\label{tab:notations}Notations used throughout the paper.}
\end{table}

\subsection{Un/Weakly-supervised Setups}

For data source, we used the standard 600K set from \citet{conceptnet-600k} (CN-600K). It is a curated version of ConceptNet 5 \citep{speer2017conceptnet}, a well-known commonsense KB, which partly originated from the Open Mind Common Sense (OMCS) list of commonsense fact sentences. Despite no explicit pairing between sentences from OMCS and paths from CN-600K, both datasets cover a related set of facts making them good candidates for our unsupervised and weakly-supervised training. ATOMIC \citep{sap2019atomic} was created from many data sources (books, n-grams, etc.) not easily accessible, leaving us with no relevant text corpora to pair with. Therefore, we only use CN-600K.

The weakly-supervised dataset was obtained by doing fuzzy matching of sentences \xA to paths \xB. Each sentence is mapped to a list of paths, and each path to a list of sentences. Note that KB and text set do not align exactly; noise and mislabelling are inherently present. Despite these constraints, we can investigate the effects of weak supervision on model performance, and vary the amount of it by changing the fraction of available paired data. 
More details about how this weakly-supervised dataset\footnote{We plan on releasing this dataset publicly.} was created are in \Sec{sec:datasets}.



\section{Model}
\label{sec:model}

Given a dataset $X$ of paths and sentences, let $x_t^k$ be its $k$-th random sample of type $t$, for $k=1, \ldots, N$ and type $t\!\in\!\{\text{A}, \text{B}\}$, where \xA is a  sentence and \xB is a path in the KG. 
Given the input $x_t^k$ to the model, the corresponding generated output will be denoted as $x_{tt^\prime}^k$, where $tt^\prime$ is the transfer direction, i.e., $tt^\prime \in \{\text{AA}, \text{AB}, \text{BB}, \text{BA}\}$. 
For example, given path $\xB^k$, $\xBA^k=\tBA(\xB^k)$ denotes the corresponding generated sentence. 
Similarly, given $x_{tt^\prime}^k$ as input, $x_{tt^\prime t^{\prime\prime}}^k$ denotes additional possible transfer directions, out of which we will be only interested in $tt^\prime t^{\prime\prime}\in\{\text{ABA}, \text{BAB}\}$, as they represent the \emph{back-translations} of a sample from a type $t$ back to itself, since $t^{\prime\prime}=t$.
Given input sentence $\xBA^k$, $\xBAB^k$ denotes its generated back-translation such that $\xBAB^k\!=\!\tAB(\xBA^k)\!=\!\tAB(\tBA(\xB^k))$. 
A model with perfect generation and back-translation would yield $\xBAB^k=\xB^k$ as $\tAB(\tBA(x))$ would be the identity function.
Note that to reduce clutter, we drop the dataset index $k$ from the notations.
\subsection{Losses}
\label{sec:losses}
There are three types of losses we employ in our training. 
The reconstruction loss is defined as 
\begin{align*}
\mathcal{L}_{\textsc{rec}} \!=\!\!\!\!\! \underset{\xA \sim X}{\mathbb{E}}\left[\minus\log p_{\textsc{aa}}(\xA)\right] + \!\!\! \underset{\xB \sim X}{\mathbb{E}}\left[\minus\log p_{\textsc{bb}}(\xB)\right],
\end{align*}
where $p_{\textsc{AA}}(\xA)$ is the distribution for the reconstructed sentences $\xAA=\tAA(\xA)$, and $p_{\textsc{BB}}(\xB)$ for paths $\xBB=\tBB(\xB)$. 
To enable model to perform transfers, we also employ the so-called back-translation, or cycle loss: 
\begin{align}
    \mathcal{L}_{\textsc{bt}} &= \underset{\xA \sim X}{\mathbb{E}}\left[\minus\log p_{\textsc{aba}}(\xA| \xAB)\right]\nonumber\\
    &+\underset{\xB \sim X}{\mathbb{E}}\left[\minus\log p_{\textsc{bab}}(\xB|\xBA)\right].    
\label{eq:loss_bt}    
\end{align}
Unsupervised training minimizes the combined loss $\mathcal{L}_{\textsc{rec}}+\mathcal{L}_{\textsc{bt}}$.
When supervised data is available, we can additionally impose a supervision loss for the paired data:
\begin{align}
    \mathcal{L}_{\textsc{sup}} &= \underset{\xA, \xB \sim X}{\mathbb{E}}\left[\minus\log p_\textsc{ab}(\xB|\xA)\right] \nonumber\\
    &+\underset{\xA, \xB \sim X}{\mathbb{E}}\left[\minus\log p_\textsc{ba}(\xA|\xB)\right].  
\label{eq:loss_sup}
\end{align}
Supervised training minimizes the combined loss $\mathcal{L}_{\textsc{rec}}+\mathcal{L}_{\textsc{bt}}+\mathcal{L}_{\textsc{sup}}$.
We explore the impact of $\mathcal{L}_{\textsc{rec}}$, $\mathcal{L}_{\textsc{bt}}$, and $\mathcal{L}_{\textsc{sup}}$ on training with an ablation study detailed in \Sec{sec:expts_loss}.
Note that when the supervised data is limited, i.e., the sentence-path correspondence is only known for some pairs $(\xA, \xB)$, the loss in \eqref{eq:loss_sup} is adjusted accordingly and averaged only over the available pairs. As we show in \Sec{sec:supervision}, this modification enables us to examine the effect of the amount of supervision on the model performance.

\section{Experimental Setup}
\label{sec:train}

In this Section, we provide details about model architecture, training procedure, and data processing.

\noindent\textbf{Model Architectures}. We explored several architectures for our encoder and decoder: GRU \citep{chungGRU}, Transformer \citep{vaswani2017attention}, and BERT \citep{devlin2019bert}. 

All possible pairings of encoder and decoder, provide a total of nine different combinations. 
Some of them, such as Transformer-BERT, GRU-BERT, or BERT-BERT are not valid, since BERT can only serve as an encoder model. 
Other configurations were rejected based on their performance on our validation set. 
In particular, we found that GRU-Transformer and BERT-Transformer just did not perform well, setting them aside. 
In the end, we selected the following three combinations: 
(1) GRU-GRU, since GRU is a simple, well-known RNN architecture, relatively easy and fast to train; 
(2) BERT-GRU, selected to leverage potentially better encoding representation from BERT, a well-known encoder;
(3) Transformer-Transformer (or Trans-Trans), chosen to explore another seq2seq architecture.
All models parameters are trained, while for BERT we fine-tune the pre-trained model.

\noindent\textbf{Teacher Forcing}. During training, in the recurrent generation steps, we employ teacher forcing by feeding the ground truth tokens only 20\% of the time, using the model output otherwise. We found this strategy very effective at preventing overfitting and increasing the overall generation quality. 

\noindent\textbf{Back-Translation}. Traditionally, when using the back-translation loss in \eqref{eq:loss_bt}, gradients are back-propagated through the encoder-decoder structure twice to reflect the two transfers being done (e.g., $\xA \rarrow \xAB \rarrow \xABA$). 
However, we observed better training behavior and model performance by detaching the gradients after the first pass through the system.
The model still sees both back-propagations but the training becomes more stable.
 
\noindent\textbf{Data processing.} In our dataset, a path such as  $(e_h, r, e_t)$ = ("yeast", "is a", "ingredient in bread") will be encoded as "\sep yeast \sep is a \sep ingredient in bread \sep{}", including the special token \sep, as done in \citet{yao2019kgbert}.
Similar to \citet{devlin2019bert}, we mask a token by replacing it with a special token \mask. For a path, we replace either the head/tail by the mask token 50\% of the time, e.g., $("\mask{}", r, e_t)$ for head masking. For text, we replace a randomly picked token with the mask token 10\% of the time. 
For more robust training, we mask tokens in both text and paths input sequences and reconstruct the original. This is in the same spirit as recently proposed masked language modeling techniques \cite{devlin2019bert,liu2019roberta}. While this technique helps learning robust encoder-decoder, it also inherently prepares the model for the link-predictions of KB completion evaluation.

\section{Results and Discussions}
\label{sec:expts}
In this Section we discuss in detail our dataset construction, evaluate the proposed models, and compare performances to existing baselines.

\begin{table*}[ht!]
\centering
\begin{tabular}{lcccccc}
  Conventional KB Completion & MRR & HITS@1 & HITS@3 & HITS@10 & GED$\downarrow$\\
  \midrule
  \textsc{DistMult} \citep{yang2015embedding}      &  8.97   &  4.51   &  9.76   & 17.44  &  - \\
  \textsc{ComplEx} \citep{trouillon2016complex}    & 11.40   &  7.42   & 12.45   & 19.01  &  - \\
  \textsc{ConvE} \citep{dettmers2018convolutional} & 20.88   & 13.97   & 22.91   & 34.02  &  - \\
  \textsc{ConvTransE} \citep{shang2019end}         & 18.68  & 7.87  & 23.87  & 38.95  &  - \\
  S+G+B+C \citep{malaviya2020commonsense}          & 51.11  & 39.42 & 59.58 & 73.59 & - \\  
  \midrule
  Generative KB Completion & & & & & \\
  \midrule
  \textsc{DualTKB}$_{\text{GRU-GRU}}$, $\rho=0.5$ & \textbf{63.10} & \textbf{55.38}	 &	\textbf{69.75} & \textbf{74.58} & 12.5	 \\
  \textsc{DualTKB}$_{\text{BERT-GRU}}$, $\rho=0.2$ & 61.32  & 53.79 & 67.62 & 72.29  & 12.0 \\
  \textsc{DualTKB}$_{\text{Trans-Trans}}$, $\rho=0.5$ & 50.54 &	44.54 &	55.12 &	59.67 &	10.0 \\
  \cmidrule(lr){2-6}
  \textsc{DualTKB}$^*_{\text{GRU-GRU}}$, $\rho=0.5$ & 50.87 & 44.58	 &	55.46 & 60.12 & 9.0	 \\
  \textsc{DualTKB}$^*_{\text{BERT-GRU}}$, $\rho=0.5$ & 57.79  & 50.25 & 63.75 & 69.54  & 11.0 \\
  \textsc{DualTKB}$^*_{\text{Trans-Trans}}$, $\rho=1.0$ & 40.93 &	35.67 &	44.38 &	48.79 &	\textbf{8.0} \\
  
\bottomrule
\end{tabular}
\caption{Conventional and generative KB completion results for ConceptNet test set for (filtered) MRR, HITS and GED metrics. Models in training w/ best MRR evaluations were selected for testing. Models with asterisk $*$ were selected based on best BLEU2 score for BA text generation task. Supervision ratio is indicated as $\rho$. Results from conventional KB completion methods quoted at the top of the table are as reported in their respective papers.}
\label{tab:kg_completion}
\end{table*}

\begin{table}[ht]
\begin{tabular}{lcccc}
\toprule
   & B$_2$   & B$_3$   & R$_L$   & B$_{F1}$  \\
\cmidrule(lr){2-5}
\textsc{DualTKB}$_{\text{GRU-GRU}}$    &  0.32 &       0.24 &        0.46 &               0.89     \\
\textsc{DualTKB}$_{\text{BERT-GRU}}$    &  0.32 &       0.25 &        0.46 &               0.88     \\
\textsc{DualTKB}$_{\text{Trans-Trans}}$ &  0.45 &       0.37 &        0.56 &               0.91     \\
\cmidrule(lr){2-5}
\textsc{DualTKB}$^*_{\text{GRU-GRU}}$     &  \textbf{0.49}  &       \textbf{0.42} &        \textbf{0.61} &               \textbf{0.92}     \\
\textsc{DualTKB}$^*_{\text{BERT-GRU}}$    &  0.37 &       0.30 &        0.51 &               0.89     \\
\textsc{DualTKB}$^*_{\text{Trans-Trans}}$ &  0.47 &       0.39 &        0.57 &               0.91     \\
\bottomrule
\end{tabular}
\caption{BA text generation evaluation results for BLEU2~(B$_2$), BLEU3~(B$_3$), RougeL~(R$_L$), and F1 BERT-score (B$_{F1}$). Models correspond to the ones in \Tab{tab:kg_completion}. Models with $*$ are selected by best B2 scores.}
\label{tab:BA_results}
\end{table}

\begin{table*}[ht]
\centering
\begin{tabular}{lcccccc}
\toprule
Losses    & \multicolumn{2}{c}{DualTKB$_{\text{GRU-GRU}}$} & \multicolumn{2}{c}{DualTKB$_{\text{BERT-GRU}}$} & \multicolumn{2}{c}{DualTKB$_{\text{Trans-Trans}}$} \\
\cmidrule(lr){1-1} \cmidrule(lr){2-3} \cmidrule(lr){4-5} \cmidrule(lr){6-7} 
   & MRR         & BLEU2        & MRR         & BLEU2        & MRR        & BLEU2        \\
$\mathcal{L}_{\textsc{rec}}$ + $\mathcal{L}_{\textsc{bt}}$ + $\mathcal{L}_{\textsc{sup}}$ & 63.10    &    0.32    &  61.32     &  0.32      &  50.53     &   0.45   \\
$\mathcal{L}_{\textsc{bt}}$ + $\mathcal{L}_{\textsc{sup}}$     &    17.09    &     0.48  &    0.14    &     0.07  &    45.42   &     0.46  \\
$\mathcal{L}_{\textsc{bt}}$ + $\mathcal{L}_{\textsc{rec}}$     &    20.08    &     0.25  &    52.52    &     0.03  &    23.56   &     0.26  \\
$\mathcal{L}_{\textsc{rec}}$ + $\mathcal{L}_{\textsc{sup}}$    &    46.16    &     0.46  &    57.57    &     0.34  &    44.08   &     0.42  \\
\bottomrule
\end{tabular}
\caption{Ablation study on different set of losses across various models}
\label{tab:ablation}
\end{table*}

\subsection{Datasets}
\label{sec:datasets}

We designed our own dataset due to the lack of available resources of paired text corpus and commonsense KBs. 
For KB, we started from CN-600K \citep{conceptnet-600k} derived from ConceptNet \citep{speer2017conceptnet}. 
As mentioned earlier, we define a path as composed of two nodes and a directed edge connecting them. 
We only keep paths from CN-600K with confidence score greater than 1.6 for a total of 100K edges. Sentences in our dataset come directly from OMCS free-text sentences, removing those with detectable profanity. 

Note that as is, this dataset can only be used for unsupervised training since its path-sentence pairs are not matched. To enable weak supervision, we map paths to sentences (and sentences to paths) using fuzzy matching directly on their sequence of tokens. Fuzzy matching uses term-frequency (tf) and inverse document frequency (idf) with n-grams to compute a cosine similarity score between two sets of strings \citep{fuzzymatch}. 

Only the top matching pair is kept if its similarity score is greater than 0.6 to ensure a minimum quality for our matches. This score of 0.6 (from a range of -1 to 1) was chosen empirically after noticing that the path-sentence pairs quality degraded quickly for lower scores, as some sentences did not have a good match. Indeed, \emph{not all} OMCS sentences were used to create ConceptNet.

We create dataset splits \emph{train} (240K) and \emph{dev} (10K) under the strict policy that no path nor sentence in dev can exist in train, to ensure validation on text and path unseen in training.

During training, we can vary the amount of weak supervision available by applying the supervision loss \Lsup in \eqref{eq:loss_sup} to a subset of the data. Note that for the pairs where \Lsup is not applied, the training becomes unsupervised since the remaining losses $\Lrec+\Lbt$ cannot exploit the presence of a match. Therefore, by changing the size of the subset to which \Lsup is applied we can go from full weak-supervision to unsupervised training.

Evaluations are reported on 1200 positive tuples of test set from the original split in \citet{conceptnet-600k} as done in prior works on commonsense KB \cite{malaviya2020commonsense}. Irrespective of the choice of encoder, we tokenize the entire dataset (paths and sentences) using the BERT tokenizer.

\subsection{Metrics and Results}
The challenge in evaluating our models is that they accomplish different generation tasks, each with its own adequate metrics. 
Detecting an overall best model is therefore ambiguous and should be potentially application specific. 
Nevertheless, we present a set of metrics and results for all the important generation tasks for three model encoder-decoder architectures: GRU-GRU, BERT-GRU, and Trans-Trans, as well as discuss the model selection process.
Results are obtained over 40 epochs of training. We present both the best results for each metric, and their averages over 4 models trained with distinct random seeds in Appendix in \Tab{tab:4seeds}. 
Note that we refer to a generation task simply by its direction, i.e., use ABA for \tABA.

\noindent\textbf{AA, ABA:}
The performance on reconstruction task AA is indicative of the quality of autoencoding, while the back-translation task ABA, besides reconstruction also helps in evaluating implicitly the quality of transfer. We use the following metrics to evaluate generated sentence quality: BLEU2 (B2), BLEU3 (B3), ROUGEL (R$_L$), and F1 BERT-scoring metric (B$_{F1}$) \cite{zhang2019bertscore} in \Tab{tab:text_result_supp} in Appendix. We observed that reconstruction AA is almost perfect in all metrics, while ABA is lagging as it must do first the translation and then recover back the sentence, a more challenging task.
GRU-GRU is not as good for ABA as for AA, while Trans-Trans provides strong ABA results. BERT-GRU is ahead of the pack for ABA. Overall the models can handle AA and ABA reasonably well.

\noindent\textbf{B$_m$B: KB Completion:}
KB completion is a common task for comparing models. 
Previous work established link prediction as the standard for evaluation relying on MRR (Mean Reciprocal Rank) and HITS as evaluation ranking metrics \citep{yang2015embedding, dettmers2018convolutional, bordes2013translating}.
For tail prediction, $e_t$ in all test tuples $(e_h, r, e_t)$ are replaced by other nodes $\tilde{e}_t$ to produce corrupted tuples.
A validity score is evaluated for all new corrupted tuples $(e_h, r, \tilde{e}_t)$, and a ranking is provided for the tuple with
ground truth $e_t$. 
Head prediction is done similarly, often by reversing the relation $r$ to $r^{-}$ and ranking $e_h$ given $(e_t, r^{-})$.
As in \citet{bordes2013translating}, filtered versions of MRR and HITS are preferred as they filter out corrupted tuples present in training, dev, and test sets. We use filtered MRR and HITS for all results in this paper.
We are aware of the recent critic of this KB completion evaluation by \citet{akrami2020realistic}. 
However, since the community is yet to establish another methodology, and to be able to relate to previous work, we follow this procedure, albeit with a twist.
For commonsense KB, \citet{malaviya2020commonsense} mention the difficulty of repurposing generative models to ranking tuples for link-prediction.
Our model is trained to generate a path tuple, not evaluate the quality score of its validity. 
Entities, relationship are all sequences of tokens that need to be generated properly, which does not fit well in the \emph{conventional} KB completion evaluation framework.
Therefore, we define a new meaningful commonsense KB completion task for generative models, and present the challenges that arise from it.

Since our models are trained from masked inputs, they can easily generate a new path $\hxB=\tBmB(\xB^m)$ from masked tuples $\xB^m=(e_h,r,.)$, or $\xB^m=(.,r,e_t)$.
This B$^m$B generation task can be done from any masked tuple, allowing for tail and head predictions without resorting to inverting the relationship $r$. 
For both predictions, the model generates a \emph{complete} edge, not just $e_h$ or $e_t$ as for generative model COMET \citep{bosselut2019comet}.
However, the generated edge may not be a proper sequence of tokens for a $(e_h,r,e_t)$ tuple, especially in early stages of training where the model is still learning the tuple structure.
For improper generated tuples, tail and head predictions become impossible as the generated sequences of tokens  cannot parse into correctly-formed tuples. 
For these cases, a worst case ranking score is given since no meaningful ranking is achievable, while their rankings are still required by MRR and HITS metrics.
Our results are therefore \emph{very} sensitive to the correct form of generated paths.

To summarize, for commonsense KB
completion, we follow these steps: 
(1) Provide an input triple w/ masked entity (tail/head); 
(2) Generate a complete triple from our model (including a candidate for the missing/masked entity);
(3) Score this predicted triple along with all corrupt triples (where the masked entity was replaced with all possible node candidates from our data) using fuzzy matching against the ground truth triple;
(4) Remove corrupt triples before scoring if they already exist in train, test, or dev (filtering).

Defining a scoring function brings also another challenge.
Indeed, even when producing proper tuples from masked entries, our generative model can still change some tokens in the unmasked part of the input tuple, i.e., for $\xB^m=(e_h,r,.)$, we could get a generated $\hxB=\tBmB(
\xB^m)$ with $\hxB=(\hat{e}_h, \hat{r}, \hat{e}_t)$, where $e_h \neq \hat{e}_h$ and/or $r \neq \hat{r}$.
There is no guarantee that the original pair $(e_h,r)$ remains identical. 
The solution of using the original $(e_h,r)$ for scoring, regardless of $(\hat{e}_h, \hat{r})$, is not fair as it will provide better performance than really achieved by the generative model.
Therefore, for scoring function, we decided to compute the ranking scores using fuzzy matching of corrupted tuples, using the \emph{whole} generated tuple sequence, taking into account any potential flawed generation of $(\hat{e}_h, \hat{r})$.
MRR and HITS computation is exactly the same as in conventional KB completion.

We provide results for our KB completion in \Tab{tab:kg_completion} where we added results from previous related work.
During training, if models with the best MRR evaluation results are selected and used for testing, then GRU-GRU shows overall better performance. However, if BLEU2 metric is used for the selection, then BERT-GRU achieves higher scores. We observed that the overall performance of Trans-Trans was behind in these evaluation metrics.
It is to be expected that not all metrics are strongly correlated and peak at the same time during training. 
It also confirms that for model selection, a weighted composite of all metrics would be a better approach.

\noindent\textbf{BA: Sentence evaluation:}
The generation of sentence from path is another important task. For qualitative results, examples of such generation are shown in \Fig{fig:gentext} and in Appendix~\ref{sec:app_graph_gen}.   
For quantitative results, we evaluated the sentences using B2, B3, R$_L$, and B$_{F1}$ metrics, as reported in \Tab{tab:BA_results} for the same set of models as in \Tab{tab:kg_completion}. We compute these metrics against our weakly-supervised data, created using fuzzy matching.
Selecting models solely on MRR has its shortcomings for other tasks. Models selected on MRR have relatively decent performances on KB completion metrics, but lag behind when compared to the models selected under text evaluation metrics. GRU-GRU and Trans-Trans are particularly good at this task, while BERT-GRU is in third place.

\begin{figure}[t]
\centering
\includegraphics[width=\columnwidth]{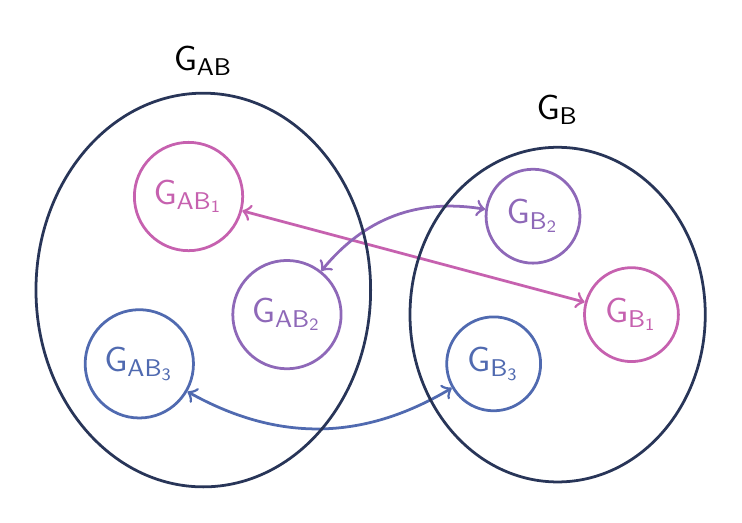}
\caption{GED computation based on local subgraphs. GED is computed as an average of corresponding local graphs. As the size of local graph increases, the computed value better approximates the global GED. In our experiments we used subgraphs consisting of 10 paths due to computational cost. Larger numbers of paths result in prohibitive GED computation cost.}
\label{fig:ged}
\end{figure}

\noindent\textbf{AB: Graph Edit Distance:}
In contrast to the single-instance evaluation, examining each generated path independently from others, we propose now to look at the generated graph as a whole and compare it to the ground truth one (based on our weakly-supervised dataset). 
In other words, given input sentences \xA, we generate the corresponding paths \xAB and compare them to the ground truth \xB. 
One of the metrics to compute graph similarity is Graph Edit Distance (GED) \citep{chen2019efficient}, which finds minimum cost path (consisting of node/edge substitutions, deletions, and insertions) transforming one graph into another. 
Since in general exact GED computation is NP-hard and practically infeasible for large graphs, we propose an approximation based on local subgraphs, as illustrated in \Fig{fig:ged}. 
To define the cost of matching nodes (corresponding to heads and tails) and arcs (corresponding to relationship operators), we encode them into feature vectors using BERT and compare the value of their euclidean distance to a predefined threshold to identify matching nodes/edges. 
GED values are reported in \Tab{tab:kg_completion} for all of our models. Trans-Trans has the lowest GED for all cases. 

\subsection{Ablation Study of Losses}
\label{sec:expts_loss}
In our ablation study, we evaluate the effect of each loss \Lrec, \Lbt, and \Lsup on the overall model performance. As seen in \Tab{tab:ablation}, by removing any of the losses, the MRR/HITS performance drops compared to the full loss case. At the same time, for different models, each loss has its own impact: e.g., \Lbt has more value for GRU-GRU model, while the availability of weak supervision \Lsup is more important for Trans-Trans architecture. Another interesting conclusion is that although \Lsup explicitly and \Lbt implicitly both control the quality of transfers (\tAB and \tBA), they remain complementary to each other, i.e., there is still a benefit of using both in the same loss.

\subsection{Impact of Supervision Ratio}
\label{sec:supervision}
\begin{figure}[t]
\includegraphics[width=\columnwidth]{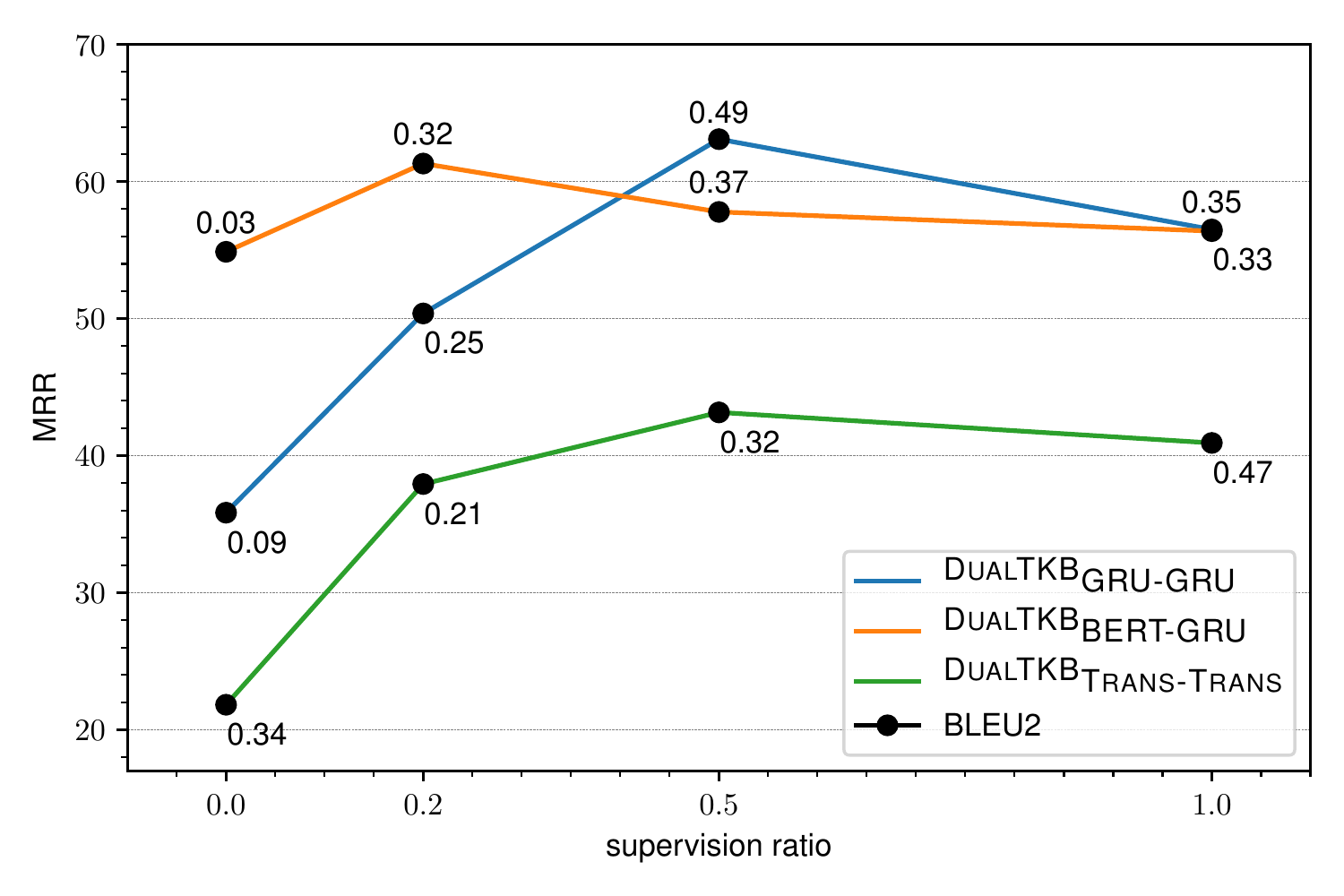}
\caption{Models MRR and BLEU2 performance for supervision ratio 0, .2, .8 and 1.0. After gains from slight supervision, full supervision degrades performances, except for BLEU2 for Trans-Trans models.}
\label{fig:supervision}
\end{figure}

The impact of weak-supervision on model performance is illustrated in \Fig{fig:supervision}.
For all the models, even a slight supervision amount leads to a rapid improvement for both MRR and BLEU2 metrics, which
vindicates our decision to use weak-supervision.
Importantly, model performance tends to peak before reaching full supervision: only our Trans-Trans model still sees an improvement for B2 scores at full supervision, while both GRU-GRU and BERT-GRU trained at 50\% weak-supervision lead to better models.
This can be explained by the nature of weak supervision itself. 
Fuzzy matching relies only on tf-idf and n-grams, which is not perfect when dealing with paths and sentences with different grammar structures. This procedure is inherently noisy and can pair sentences and paths with related but, nevertheless, not exact semantic matches.  The full weak-supervision usually brings noise into the training, thus harming it more than helping. This conclusion is confirmed by the lower performance of our models at full supervision.

\section{Conclusion}

In this paper we proposed to use a dual learning bridge between text and commonsense KB. In this approach, a generative model is trained to transfer a sentence to a path and back. Assembling paths together results in a graph showing the presence of inherent structure, while generated sentences exhibit coherent and relevant semantics. For evaluation, we proposed a novel commonsense KB completion task tailored to generative models. 
Although our model is designed to work in unsupervised settings, we investigated the impact of weak-supervision by creating a weakly-supervised dataset and showed that 
even a slight amount of weak-supervision improves significantly model performance.
The current work is one step towards the overarching goal of KB construction/completion \emph{and} generation of human-readable text from KBs. Future work can focus on expanding the capabilities to generating whole paragraphs of text from graphs in KB, as well as converting large parts of text into coherent graph structures.

%

\bibliography{emnlp2020}
\bibliographystyle{acl_natbib}


\appendix
\section{Supplementary}
\label{sec:suppl}

\subsection{Experimental Setup and Training}
\label{sec:setup_suppl}

All our models were built using an NVIDIA V100 GPU, while validation was done with a CPU-only setup (using 2 cores).
Each training epoch (240K samples) took between 1 hour for GRU-GRU (fastest), just over 1 hour for BERT-GRU, and 1.5h for Trans-Trans models (slowest).
Our validation (10K samples) takes about 1.5 hours to evaluate as we need 2 passes for MRR and HITS (head and tail predictions), as well as a third pass for all NLP metrics for AA, ABA, and BA.
Evaluation on our test set (1.2K samples) takes a matter of about 6 minutes for each head and tail prediction for MRR and HITS, with a third pass for NLP metrics, for a total of about 20 minutes. BLEU2 evaluation is relatively fast. GED evaluation can take up to 40 minutes.
During training, we evaluate models at the end of each epoch, and use the best model over 40 epochs for testing.

All our models were built using PyTorch.
They are trained with a batch size of 32 for a supervision ratio of 0.5. Masking for sentences is performed with a probability of 10\% for each token while masking for paths tuples entities are set at 50\% (when selected, all tokens of an entity are masked).

GRU models are trained with a learning rate of either $10^{-3}$ or $5\times10^{-3}$ ($10^{-4}$ yielded poor results) 
Transformer models were trained with a predefined learning rate schedule called "NoamOpt" providing a warm-up phase up to 20K training minibatches before a slow exponential decay. All our Transformer models had three sub-layers with three heads in the multi-head attention, while 
GRU models had a hidden size of 100 with a single recurrent layer.

All trainings had a seeded random number generator to ensure repeatability.
For instance, results from \Tab{tab:text_result_supp} were obtained for one common seed for all models.
For  results in \Tab{tab:4seeds}, 4 distinct seeds were used to show performance based on the average of the 4 individual model performances (standard deviation is also provided).

MRR and HITS metrics are defined in details in \citet{dettmers2018convolutional} -- code associated with the paper provides implementations in Python. We reimplemented them for better integration in our own codebase. BLEU metrics were used from NLTK implementations.

When running multiple supervision ratio for weak-supervision, hyper-parameters were all fixed to the values provided above, only the supervision ratio was changed.

\subsection{AA, ABA: Reconstruction and Back-Translation Tasks}

\Tab{tab:text_result_supp} show results for our models GRU-GRU, GRU-BERT, and Trans-Trans for BLEU2, BLEU3, ROUGE, and F1 BERT score for both AA and ABA tasks. AA is an easier task with excellent while ABA is more difficult. GRU-GRU and Trans-Trans excel at both AA and ABA. 

\begin{table*}[t]
\centering
\begin{tabular}{lcccccccc}
\toprule
   & \multicolumn{4}{c}{AA} & \multicolumn{4}{c}{ABA} \\
   & BLEU2   & BLEU3   & R$_L$   & BERT$_{F1}$  & BLEU2   & BLEU3   & R$_L$   & BERT$_{F1}$  \\
\cmidrule(lr){2-5} \cmidrule(lr){6-9}
    \textsc{DualTKB}$_{\text{GRU-GRU}}$     &   0.97 &  0.96  & 0.98 & 1.00 &   0.54 &  0.48 &  0.69 &  0.93 \\ 
    \textsc{DualTKB}$_{\text{BERT-GRU}}$    &   0.71 &  0.64 & 0.74 & 0.93 &   0.53 &  0.45 &  0.63 &  0.91 \\
    \textsc{DualTKB}$_{\text{Trans-Trans}}$ &   0.95 &  0.94 & 0.96 & 0.99 &   0.57 &  0.50 &  0.69 &  0.92 \\
\bottomrule
\end{tabular}
\caption{Results for text generation AB and ABA evaluation for BLEU2~(B$_2$), BLEU3~(B$_3$), Rouge-L~(R$_L$), BERT Score BERT$_{F1}$}
\label{tab:text_result_supp}
\end{table*}

\subsection{Results for Multiple Random Generation Seeds}

In \Tab{tab:4seeds}, we present results for our three model architectures using MRR-HITS and BLEU metrics as  averages over 4 models for each architecture built with distinct random generator seeds, trained with our default hyper-parameters described in \Sec{sec:setup_suppl}. We report average and standard deviation for every metrics.

\begin{table*}[t!]
\centering
\begin{tabular}{lcccccc}
\toprule
     & MRR & HITS@1 & Hits@3 & HITS@10 & BLEU2 & BLEU3\\
\textsc{DualTKB}$_{\text{GRU-GRU}}$      &  51.13$\pm$5.59   & 44.78$\pm$5.12       & 55.89$\pm$6.38       &  60.68$\pm$6.38       &    0.52$\pm$0.07         & 0.41$\pm$0.08       \\
\textsc{DualTKB}$_{\text{BERT-GRU}}$     &  57.49$\pm$4.29   & 50.05$\pm$3.80       &  63.59$\pm$4.95      & 68.57$\pm$4.90        &  0.46$\pm$0.02 & 0.34$\pm$0.01             \\
\textsc{DualTKB}$_{\text{Trans-Trans}}$  &  43.87$\pm$4.65   & 37.57$\pm$4.93       &    47.80$\pm$5.02    &  54.02$\pm$4.11       & 0.48$\pm$0.08 & 0.38$\pm$0.05            \\
\bottomrule
\end{tabular}
\caption{Results (Mean$\pm$SD) of various models across 4 different seeds for our random number generator. MRR, HITS metrics corresponds for KB completion task, whereas BLEU scores are shown for BA text generation task}
\label{tab:4seeds}
\end{table*}

\subsection{AB: Graph Generation}
\label{sec:app_graph_gen}

\begin{figure}[ht]
\centering
\includegraphics[width=\columnwidth]{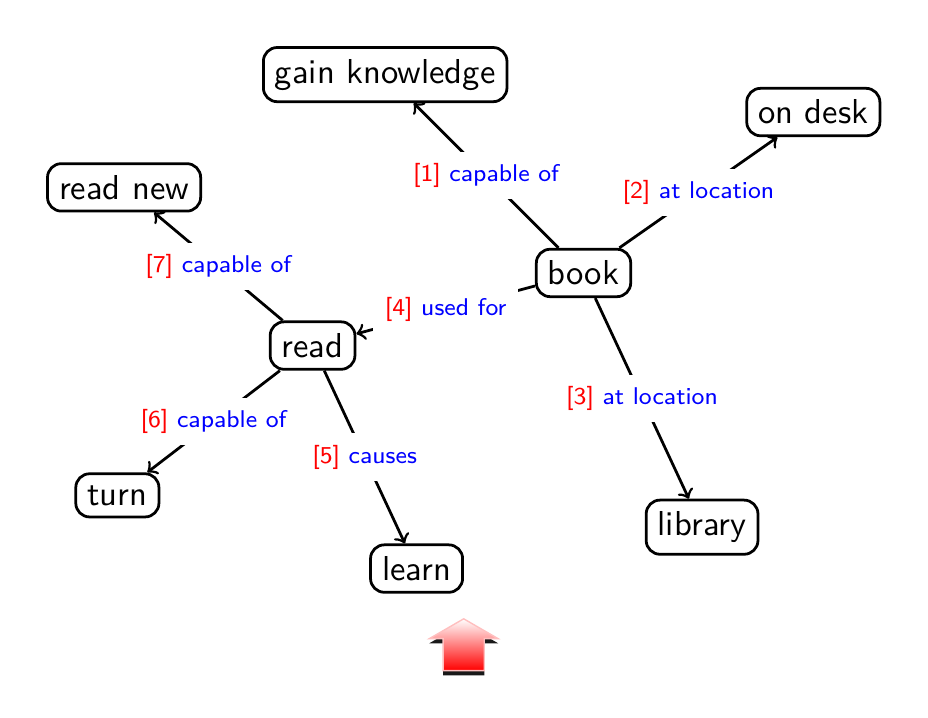}
{
\centering
\resizebox{\columnwidth}{!}{
\begin{tabular}{clcl}
    & & & \\
    {[}1{]} & book can store knowledge &
    {[}2{]} & a something you find on your desk is a book \\
    {[}3{]} & a book is part of library &
    {[}4{]} & book is to read \\
    {[}5{]} & to read is to learn &
    {[}6{]} & a reader can turn a page \\
    {[}7{]} & you can read the news in a newspaper & 
\end{tabular}
}
}
\caption{Text to Path. A part of a larger graph generated by our system based on the test split of ConceptNet dataset. The shown sentences are a subset of the inputs provided to the model.}
\label{fig:gengraph2}
\end{figure}

\begin{figure}[ht!]
\centering
\includegraphics[width=\columnwidth]{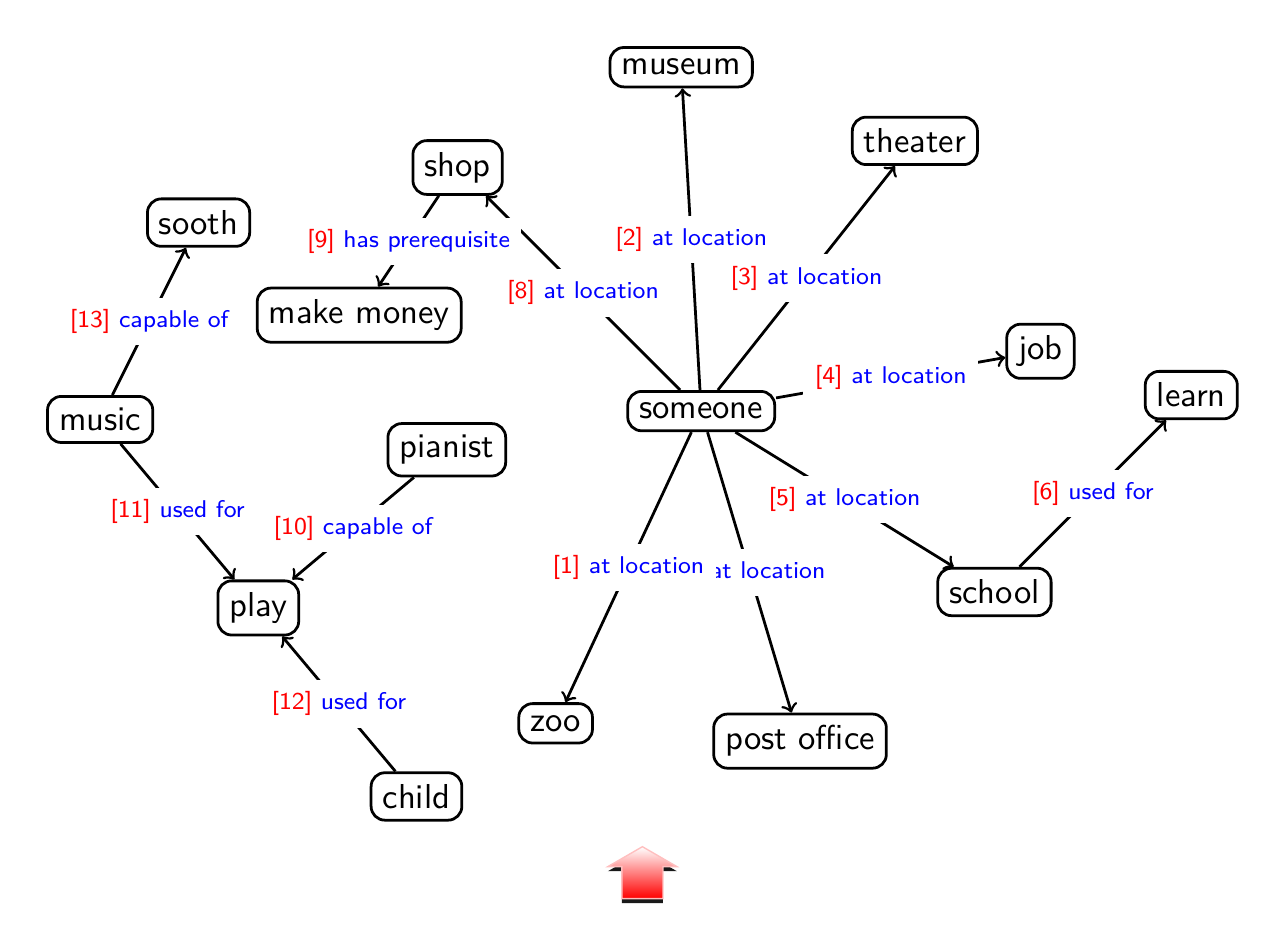}
{
\centering
\resizebox{\columnwidth}{!}{
\begin{tabular}{clcl}
    & & & \\
    {[}1{]} & something you find at a zoo is an animal &
    {[}2{]} & someone can be at museum \\
    {[}3{]} & someone can be at the theater &
    {[}4{]} & someone can be at a job interview  \\
    {[}5{]} & something can be at school &
    {[}6{]} & in school , you can learn \\
    {[}7{]} & someone can be at a post office & 
    {[}8{]} & someone can shop \\
    {[}9{]} & if you want to shop then you should have money & 
    {[}10{]} & pianist play piano  \\
    {[}11{]} & play music & 
    {[}12{]} & a child can play  \\
    {[}13{]} & music can soothe & 
\end{tabular}
}
}
\caption{Text to Path. A part of a larger graph generated by our system based on the test split of ConceptNet dataset. The shown sentences are a subset of the inputs provided to the model.}
\label{fig:gengraph3}
\end{figure}

\begin{figure}[t!]
\centering
\includegraphics[ width=\columnwidth]{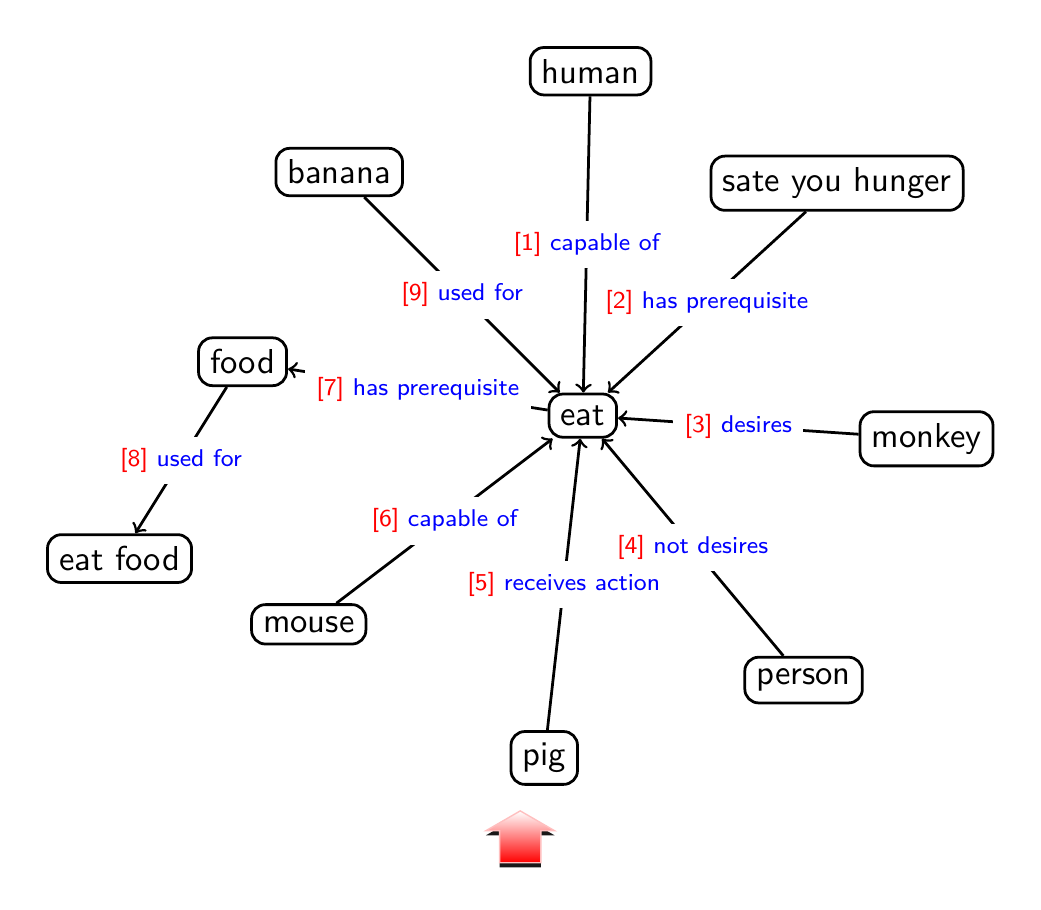}

{
\centering
\resizebox{\columnwidth}{!}{
\begin{tabular}{clcl}
    & & & \\
    {[}1{]} &  a human can eat &
    {[}2{]} & if you want to sate your hunger then you should eat \\
    {[}3{]} & monkey like to eat banana &
    {[}4{]} & a person can eat \\
    {[}5{]} & a pig can eat &
    {[}6{]} & a mouse can eat \\
    {[}7{]} & you eat food  & 
    {[}8{]} & you can use a fork to eat food \\
    {[}9{]} & banana is fruit  & 
    
\end{tabular}
}
}
\caption{Text to Path. A part of a larger graph generated by our system based on the test split of ConceptNet dataset. The shown sentences are a subset of the inputs provided to the model.}
\label{fig:gengraph4}
\end{figure}

\begin{figure}[t!]
\centering
\includegraphics[width=\columnwidth]{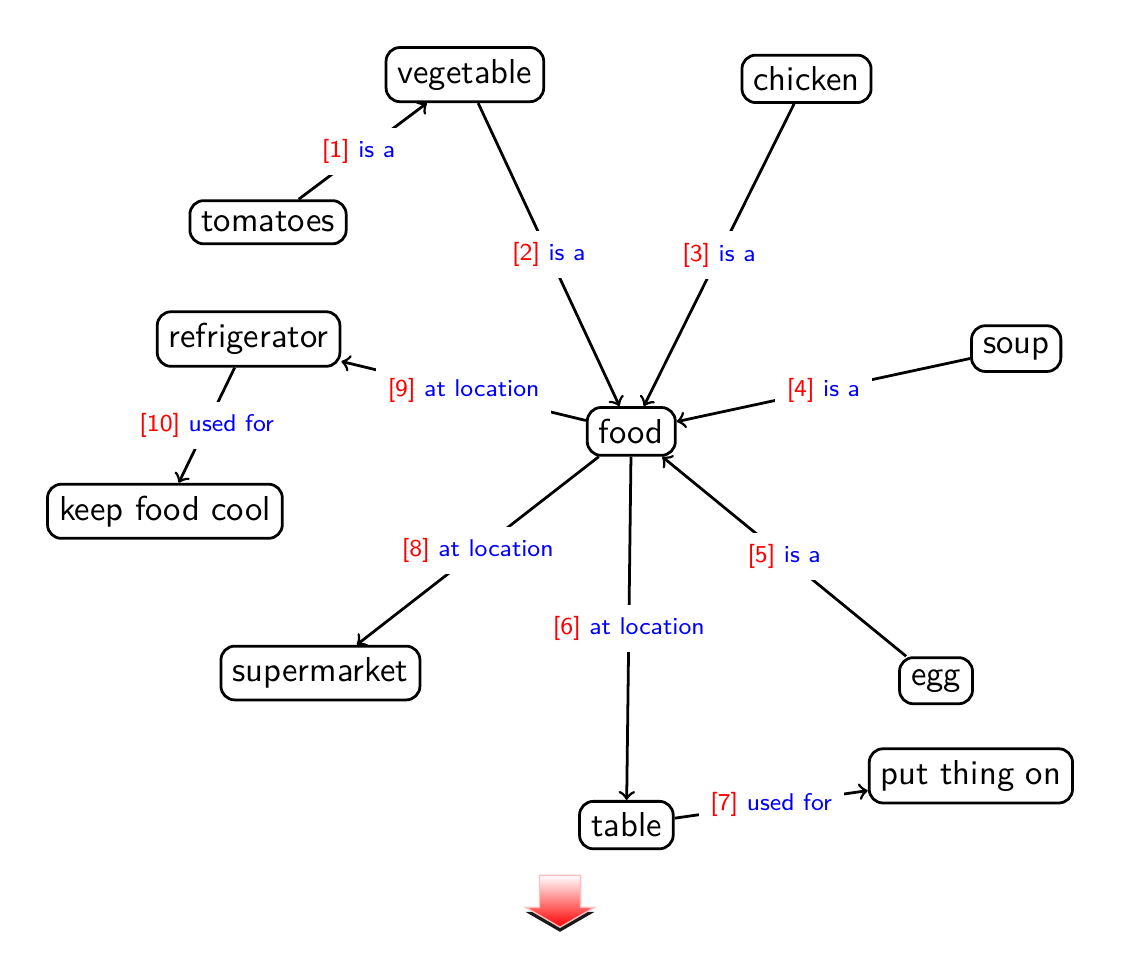}
{
\centering
\resizebox{\columnwidth}{!}{
\begin{tabular}{clcl}
    & & & \\
    {[}1{]} &  a tomatoes is a kind of vegetable &
    {[}2{]} & a vegetable is a kind of food  \\
    {[}3{]} & chicken is a kind of food &
    {[}4{]} & soup is a kind of food  \\
    {[}5{]} & an egg is a kind of food &
    {[}6{]} & something you find on table is food \\
    {[}7{]} & table is to put things on  & 
    {[}8{]} & something you find at the supermarket is food \\
    {[}9{]} & something you find in the refrigerator is food  & 
    {[}10{]} & a refrigerator is for keeping food cold \\
\end{tabular}
}
}
\caption{Path to Text. A set of sentences generated by our system from a subgrapth of the ConceptNet dataset. The paths shown in the graph are the inputs to the model.}
\label{fig:gentext2}
\end{figure}

In this Section we present additional examples of the text to path transfers, see Figures \ref{fig:gengraph2}, \ref{fig:gengraph3} and \ref{fig:gengraph4}, as well as the reverse transfer from path to text, see Figure \ref{fig:gentext2}.

\end{document}